\documentclass[letterpaper]{article} 
\usepackage{aaai2026}  
\usepackage{times}  
\usepackage{helvet}  
\usepackage{courier}  
\usepackage[hyphens]{url}  
\usepackage{graphicx} 
\usepackage{natbib}  
\usepackage{caption} 
\usepackage{algorithm}
\usepackage{algorithmic}
\usepackage[utf8]{inputenc} 

\usepackage{newfloat}
\usepackage{multirow} 
\usepackage{listings}
\DeclareCaptionStyle{ruled}{labelfont=normalfont,labelsep=colon,strut=off} 
\floatstyle{ruled}
\newfloat{listing}{tb}{lst}{}
\floatname{listing}{Listing}
\usepackage{threeparttable}
\usepackage{array} 
\usepackage{booktabs}
\usepackage{multirow}
\usepackage{float}  
\usepackage{adjustbox}
\usepackage{caption}
\usepackage[table,xcdraw]{xcolor}
\usepackage{multirow}
\usepackage{enumitem}
\usepackage[font=small,labelfont=bf]{caption}  
\usepackage[utf8]{inputenc}
\usepackage[most]{tcolorbox} 
\usepackage{enumitem} 
\usepackage{subcaption}
\newtcolorbox{promptbox}{
  enhanced,
  colback=gray!3!white,
  colframe=gray!60!black,
  coltitle=black,
  boxrule=0.4pt,
  arc=4pt,
  outer arc=2pt,
  boxsep=3pt,
  left=4pt, right=4pt, top=2pt, bottom=2pt,
  fonttitle=\scriptsize\bfseries,
  fontupper=\scriptsize,
  title=Prompt Strategies (Zero-Shot \& CoT),
  attach boxed title to top left={xshift=5pt, yshift=-1mm},
  boxed title style={
    colback=gray!15!white,
    colframe=gray!60!black,
    arc=3pt,
    outer arc=2pt,
    boxrule=0.4pt
  },
  before=\vspace{0.2em},
  after=\vspace{0.2em}
}

\usepackage{graphicx}    

\title{Evaluating LLM Alignment on Personality Inference from Real-World Interview Data}
\title{Evaluating LLM Alignment on Personality Inference from Real-World Interview Data}

\author{
  Jianfeng Zhu\textsuperscript{\rm 1},
  Julina Maharjan\textsuperscript{\rm 1},
  Xinyu Li\textsuperscript{\rm 1},
  Karin G. Coifman\textsuperscript{\rm 2},
  Ruoming Jin\textsuperscript{\rm 1}
}
\affiliations{
  \textsuperscript{\rm 1} Department of Computer Science, Kent State University \\
  \textsuperscript{\rm 2} Department of Psychological Sciences, Kent State University \\
}

\usepackage{bibentry}

\begin{document}

\maketitle

\begin{abstract}
Large Language Models (LLMs) are increasingly deployed in roles requiring nuanced psychological understanding, such as emotional support agents, counselors, and decision-making assistants. However, their ability to interpret human personality traits, a critical aspect of such applications, remains unexplored, particularly in ecologically valid conversational settings. While prior work has simulated LLM "personas" using discrete Big Five labels on social media data, the alignment of LLMs with continuous, ground-truth personality assessments derived from natural interactions is largely unexamined.
To address this gap, we introduce a novel benchmark comprising semi-structured interview transcripts paired with validated continuous Big Five trait scores. Using this dataset, we systematically evaluate LLMs performance across three paradigms: (1) zero-shot and chain-of-thought prompting with GPT-4.1 Mini, (2) LoRA-based fine-tuning applied to both RoBERTa and Meta-LLaMA architectures. and (3) regression using static embedding from pretrained BERT and OpenAI's \texttt{text-embedding-3-small}. Our results reveal that all Pearson correlations between model predictions and ground-truth personality traits remain below 0.26, highlighting the limited alignment of current LLMs with validated psychological constructs. Chain-of-thought prompting offers minimal gains over zero-shot, suggesting that personality inference relies more on latent semantic representation than explicit reasoning. These findings underscore the challenges of aligning LLMs with complex human attributes and motivate future work on trait-specific prompting, context-aware modeling, and alignment-oriented fine-tuning.
\end{abstract}
\begin{links}
    \link{Code}{https://anonymous.4open.science/status/AAAI2026_personality-D7CB}
\end{links}
\section{Introduction}
Personality is a foundational construct in psychology, defined as relatively stable and enduring patterns of thought, feeling, and behavior that distinguish individuals from each other, \cite{allport1937personality} and \cite{bleidorn2019policy}. Among various theoretical frameworks, the Five-Factor Model (FFM) \cite{murray1936basic} has emerged as the predominant empirically supported paradigm for conceptualizing the structure of personality. This model posits five broad dimensions; Openness to Experience, Conscientiousness, Extraversion, Agreeableness, and Neuroticism, which capture a vast range of individual differences. The robustness of the model is rooted in the lexical hypothesis \cite{allport1936trait}, which suggests that the most significant aspects of human personality are encoded within the natural language that people use to describe themselves and others.\\
In the contemporary digital landscape, the explosion of user-generated text on social media platforms has created an unprecedented ecological data source, fueling the rise of Automatic Personality Prediction (APP) as a key research area at the confluence of computational linguistics and psychology \cite{de2025personality}. The ability to accurately infer personality traits from text has profound implications for a multitude of applications, including personalized human-computer interaction, adaptive marketing \cite{caldwell1998personality}, mental health screening \cite{zhu2025probing}, \cite{julina_personality_su} and social science research \cite{maharjan2025large}.\\
The recent ascendancy of Large Language Models (LLMs) such as GPT-4 (OpenAI 2023), Llama 3 \cite{touvron2023llama}, and Mistral \cite{mistral2024} represents a significant paradigm shift in Natural Language Processing (NLP). These models, pre-trained on vast web-scale corpora, have demonstrated remarkable capabilities in semantic understanding, reasoning, and text generation, transforming performance benchmarks across a wide array of downstream tasks \cite{park2023generative} and \cite{wang2023rolellm}. This has naturally spurred intense interest in applying LLMs to the highly nuanced challenge of APP. The central promise is that these models can transcend the limitations of earlier methods that relied on surface-level lexical features, and instead learn to identify the subtle, context-dependent linguistic cues that are truly indicative of an individual's personality \cite{maharjan2025psychometric}. 
\\
However, a critical examination of the existing work reveals a persistent and fundamental limitation that hinders progress: the oversimplification of the personality construct itself. Despite the sophistication of both psychological theory and modern LLM architectures, a significant portion of APP research, including many initial investigations with LLMs such as \cite{hang_jiang2023personallm} and \cite{huang2024designing}, has framed the task as a binary classification problem; categorizing an user as either "high" or "low" in extraversion, for instance. This approach creates a psychometric-to-computational translation gap where the computational task fails to represent the psychological construct it purports to measure. \\
\noindent
This work is motivated by the clear and pressing need for a more rigorous and psychometrically grounded evaluation of LLMs in the domain of personality understanding. A critical gap lies in the lack of a systematic assessments of modern LLMs on the task of predicting continuous personality scores ecologically valid, real-world text. To overcome prior limitations in trait simplification and ecological validity, we introduce a systematic evaluation of LLMs for continuous personality prediction using real-world interview data. This paper makes the following contributions:
\begin{enumerate}
\item We conduct the first systematic evaluation of LLMs for predicting continuous Big Five personality scores from  real-world, semi-structured interview transcripts with validated psychometric labels.
\item We benchmark three modeling paradigms: (1) zero-shot and chain-of-thought prompting with GPT-4.1 Mini, (2) LoRA-based fine-tuning applied to RoBERTa and Meta-LLaMA, and (3) regression using embeddings from pretrained BERT and OpenAI’s \texttt{text-embedding-3-}\\\texttt{small}.
\item Trait-wise analysis reveals consistent difficulty in modeling Agreeableness and Extraversion. Most methods show limited alignment with ground-truth, highlighting the inherent challenge of inferring latent personality traits from natural language.
\end{enumerate}
These findings offer critical insight into the current limitations of LLM-based personality inference and highlight the need for more principled alignment strategies in psychologically grounded AI systems.
\section{Related Work}
Traditional personality assessment relies on standardized self-report tools such as the Big Five Inventory \cite{john1999big} and its successor, the BFI-2 \cite{bfi2}, which produce continuous Likert-scale scores across five personality dimensions. These scores are psychometrically validated and widely adopted, yet their structure poses challenges for AI models—particularly those optimized for binary classification or limited-context understanding. Bridging this psychometric-to-computational gap is essential for developing personality-aware AI systems that are both interpretable and clinically meaningful.\\
Early research in Automatic Personality Prediction (APP) employed classical machine learning techniques (e.g., SVMs, regression) \cite{kamalesh2022personality} with handcrafted features such as LIWC categories \cite{pennebaker2001linguistic}, emotion lexicon \cite{mohammad2013nrc}, or Term Frequency \& Inverse Gravity Moment (TF-IGM). The introduction of deep learning shifted focus to transformer-based models like BERT \cite{devlin2019bert} and RoBERTa \cite{liu2019roberta}, trained on large corpora and fine-tuned on social media datasets such as myPersonality \cite{schwartz2013personality} or PANDORA \cite{gjurkovic2020pandora} .While these advances improved predictive performance, most used binary labels and non-clinical data, limiting generalizability and psychometric fidelity \cite{trull2005categorical}. \\
Recent work explores LLMs from two angles: (1) measuring the personality of LLMs themselves \cite{jiang2023evaluating, peters2024large,hang_jiang2023personallm}, and (2) leveraging LLMs to assess human personality,  \cite{serapio2023personality,peters2024large,maharjan2025psychometric,zhu2025llmsinferpersonalityreal}. While zero-shot or few-shot prompting  (including Chain-of-Thought (CoT))O has been proven to be an efficient method for all kinds of inference tasks, these methods without task-specific training have been linked as ineffective for personality prediction, especially for continuous scores, yielding weak to negligible correlations with ground-truth data \cite{maharjan2025psychometric}. A more successful approach uses LLM embeddings as features for simpler, supervised downstream classifier like a 2-layer MLP, which has been shown to significantly outperform zero-shot methods \cite{maharjan2025psychometric}. Further improvements have come from parameter-efficient tuning such as LoRA \cite{hu2022lora}, as well as knowledge distillation methods like Retrieval-Augmented Generation (RAG) by \cite{lewis2020retrieval}. \\
Despite notable methodological advances, previous studies in automatic personality prediction using large language models face three key limitations. First, the majority of existing work is based on non-clinical data sources such as social media posts or synthetic text, which lack ecological validity and do not reflect the structure or content of real-world personality assessment contexts. Second, many approaches rely on simplified binary classification schemes-e.g., predicting whether a trait is high or low-thereby neglecting the continuous, multidimensional nature of personality constructs as defined in psychometrics. This simplification limits the interpretability and practical utility in applied settings. Third, zero-shot and few-shot prompting methods have shown limited effectiveness and exhibit high sensitivity to prompt formulation, making them unstable and unreliable for robust personality inference. These limitations underscore the need for studies that incorporate real world relevant datasets, support continuous trait prediction, and benchmark multiple modeling strategies under a unified framework.\\
Our study addresses these gaps by providing a comprehensive evaluation of three LLM-based paradigms, zero-shot prompting, LoRA fine-tuning, and embedding regression on a clinically grounded dataset comprising semi-structured interviews with ground-truth Big Five scores. \\
Table~\ref{tab:related_work} compares recent LLM-based APP studies, highlighting our unique contribution: the first systematic benchmark on real-world, gold-standard continuous labels using consistent psychometric metrics.
By aligning computational modeling with psychometric rigor and clinical context, our work bridges methodological innovation with practical relevance for personality-aware AI.

\noindent
Table~\ref{tab:related_work} summarizes recent LLM-based personality recognition studies, highlighting differences in model types, data modality, and labeling schemes. 

\begin{table*}
\footnotesize
\small
\renewcommand{\arraystretch}{1.1}
\centering
\caption{A Comparative Overview of Recent LLM-based Personality Recognition Studies. This table highlights the models, methodologies, and datasets used in prior work, contextualizing the contribution of this paper, which focuses on a comprehensive evaluation using a real-world dataset with continuous scores.}
\label{tab:related_work}
\begin{tabular}{p{2.5cm} p{2.3cm} p{4.6cm} p{2.1cm} p{1.5cm} p{3.2cm}}
\toprule
\textbf{Study} & \textbf{LLM(s) Used} & \textbf{Methodology} & \textbf{Dataset} & \textbf{Label Type} & \textbf{Key Finding / Limitation} \\
\midrule
\cite{christian2021text} & BERT, RoBERTa, XLNet & Employed a multi-model deep learning architecture using pre-trained language models as feature extractors, combined with model averaging. & MyPersonality & Binary & Outperformed previous works by combining multiple data sources and models, achieving high classification accuracy (e.g., 88.5\% on Twitter data). \\
\addlinespace
\cite{ji2023chatgpt} & ChatGPT & Used prompting strategies (zero-shot, one-shot, CoT); zero-shot CoT most effective for explanation and prediction.& Essays, PANDORA & Binary & CoT improves over basic zero-shot but still trails fine-tuned SOTA. \\
\addlinespace

\cite{jiang2023evaluating} & GPT-3.5, Alpaca & Developed the Machine Personality Inventory (MPI) to evaluate LLM personality and proposed a "Personality Prompting" (P²) method to induce specific traits in LLMs. & Machine Personality Inventory (MPI) & N/A (Assessment/Induction) & Provided evidence that aligned LLMs exhibit stable, human-like personality traits. The P² method was effective in controllably inducing desired personalities. \\
\addlinespace

\cite{sirasapalli2023deep} &  BERT & Proposed a data fusion methodology, mapping the MBTI dataset to Big Five traits and combining it with other datasets to create a larger corpus for training. & Essays, myPersonality, MBTI & Binary & Data fusion technique significantly increased the amount of labeled data and improved model performance, reaching 87.89\% accuracy. \\
\addlinespace
\cite{serapio2023personality} & PaLM family (18 LLMs total) & Assessed the simulated personality of LLMs by administering validated psychometric tests (e.g., IPIP-NEO, BFI) directly to the models under various prompting configurations. & Psychometric Tests (IPIP-NEO, BFI) & N/A (Assessment) & Found that personality simulated by larger, instruction-tuned LLMs can be reliable and valid. Showed that LLM personality can be shaped via prompting. \\
\addlinespace
 \cite{hu2024llm} & LLM (unspecified) & Used LLMs to generate semantic/sentiment/linguistic augmentations for contrastive distillation. & Kaggle personalitycafe, PANDORA & Binary & LLM-augmented data improves smaller task-specific models. \\
\addlinespace
\cite{peters2024large} & GPT-4, LLaMA 2, Mistral & Evaluated LLMs using zero-shot and CoT to predict continuous scores; assessed with correlation and error metrics. & Essays & Continuous & Weak correlation with ground truth ($r < 0.30$); CoT offers minimal gain. \\
\addlinespace
\cite{wang2025evaluating} & GPT-4 & Evaluated GPT-4's ability to emulate or role-play real individuals by providing it with human personality scores and having it complete questionnaires as that person. & Human personality data & Continuous & Emulated responses showed extremely high convergent validity ($r > 0.90$) with human scores but also superior (less realistic) internal consistency. \\
\cite{maharjan2025psychometric} & RoBERTa, BERT, OpenAI & Generated LLM embeddings and trained a BiLSTM to predict continuous scores. & PANDORA & Continuous & Embedding-based models outperform zero-shot approaches. \\
\addlinespace
\textbf{This Work} & GPT-4, LLaMA 3, RoBERTa & Compares (zero-shot and CoT) prompting methods, embedding-based regression, and peft fine-tuning using LoRA on real-world interviews with continuous BFI-10 labels & \textbf{Real-World  Interview} & \textbf{Continuous} & First large-scale evaluation using ecologically valid, clinically relevant data. \\
\bottomrule
\end{tabular}
\vspace{-0.5em}
\end{table*}
However, most prior work evaluates models on synthetic or social media datasets, and often uses binary classification targets, limiting their interpretability and psychometric validity. In contrast, our study provides a systematic benchmark across three major LLM-based inference paradigms—zero-shot prompting, embedding-based regression, and LoRA fine-tuning—on a real-world dataset of clinical interviews with gold-standard continuous Big Five scores. By comparing these methods under consistent evaluation metrics, our work clarifies the practical boundaries of LLM-based personality inference and offers guidance for building psychometrically aligned AI systems. Our work provides a needed empirical benchmark of this hierarchy on a real-world dataset with continuous labels.

\section{Ecologically Valid Interview Dataset }
Most prior work on language-based personality inference relies on synthetic, social media, or crowdsourced datasets that lack ecological structure or psychometric supervision. To address these limitations, we leverage a unique dataset comprising real-world, semi-structured interview transcripts from 518 adult participants, each paired with ground-truth Big Five personality scores using the validated BFI-10 instrument. This setup allows direct evaluation of alignment between model-inferred traits and human-reported psychometric measure.\\
 Each interview session resulted in approximately 15 minutes of recorded speech per participant. The speech was 
then recorded and transcribed according to convention and recommendations. Interviewers were trained to provide standardized prompts, and participants were encouraged to speak spontaneously for up to three minutes in response to each question \cite{coifman2007repressive,coifman2010distress,coifman2016context,harvey2014emotion}.\\
Representative excerpts from a single participant are shown in (Table~\ref{tab:interview_examples}).
\begin{table}[H]
\centering
\small
\setlength{\tabcolsep}{4pt} 
\renewcommand{\arraystretch}{0.95} 
\vspace{-1em}
\caption{Example Interview Responses (Participant ID: 001)}
\label{tab:interview_examples}
\begin{tabular}{p{3cm} p{4.5cm}} 
\toprule
\textbf{Prompt} & \textbf{Response Excerpt} \\
\midrule
Daily Activities & ``Beginning of the day... I have two sons... played outside... gym...'' \\
Difficult Experience & ``Being a firefighter... challenging and amazing experience... bad experiences too...'' \\
Emotion Regulation & ``You talk to people you trust at work... my wife and I have been married...'' \\
Negative Event & ``First big incident... stressful leading into it...'' \\
Positive Event & ``First baby we delivered—early morning call... heroin case...'' \\
\bottomrule
\end{tabular}
\vspace{-1em}
\end{table}
The sample was demographically diverse and balanced by sex (275 male, 278 female), with participants ranging widely in age (M = 39.39, SD = 16.33), racial backgrounds (431 White, 122 non-White/Other), and education levels (from high school to college and above). This diversity enhances the generalizability of model evaluations and supports ecologically grounded assessments of alignment.
\noindent

\section{Experimental Design}
\paragraph{Prompt-based Inference}
To evaluate the capacity of large language models for personality inference, we employ GPT-4.1 Mini via the OpenAI API (model ID: gpt-4.1-mini-2025-04-14) \cite{openai2025gpt41mini} to predict Big Five trait scores from naturalistic interview transcripts. The model is instructed to act as a psychologically insightful agent and generate a continuous score between 1.0 and 5.0 for each of the five traits. \\
\vspace{0.3em}
\begin{figure}[h]
\centering
\begin{promptbox}
{Analyze the following response and predict the speaker's Big Five personality traits using the format below.}
\vspace{0.3em}
\textbf{Input Response:} \texttt{"\{response\}"}
\vspace{0.3em}
\\
\textbf{Zero-shot Prompt:} Return a score (1.0--5.0, including half steps) for each trait: \\
\texttt{Conscientiousness: X} \\
\texttt{Agreeableness: X} \\
\texttt{Neuroticism: X} \\
\texttt{Openness: X} \\
\texttt{Extraversion: X}
\vspace{0.3em}
\\
\textbf{Chain-of-Thought Prompt:}
\begin{enumerate}[leftmargin=1em, itemsep=1pt, topsep=1pt, parsep=0pt]
  \item Summarize the emotional tone and main themes of the response.
  \item Identify linguistic patterns or statements linked to each trait.
  \item Justify the trait scores based on the above observations.
\end{enumerate}
\vspace{0.3em}
Then output trait scores in the same structured format.
\end{promptbox}
\caption{Examples of zero-shot and chain-of-thought (CoT) prompts}
\label{fig:promptbox}
\end{figure}
\vspace{0.3em}
\noindent We implement two prompting strategies. The first is a standard zero-shot prompt that directly instructs the model to return trait scores based on the interview content \cite{kojima2023largelanguagemodelszeroshot}. The second is a chain-of-thought (CoT) prompt that encourages intermediate reasoning about emotional tone and linguistic marker prior to scoring \cite{wei2023chainofthoughtpromptingelicitsreasoning}. Both prompting strategies are executed deterministically using a temperature of 0.2 \cite{Renze_2024}. The full prompt templates are shown in Figure \ref{fig:promptbox}. This prompting-based setup allows us to evaluate the alignment between model-inferred personality traits and ground-truth psychometric scores, without any parameter tuning or gradient updates.
\paragraph{LoRA-based Adaptation}
To enhance alignment between model predictions and ground-truth Big Five personality scores, we apply Low-Rank Adaptation (LoRA; \cite{hu2022lora}) for efficient fine-tuning of transformer-based language models. Specifically, we fine-tune the \texttt{meta-llama/Llama-3.1-8B-Instruct} model \cite{llama32025} separately for each trait. LoRA modules are injected into the attention and feedforward layers, allowing for parameter-efficient adaptation while preserving the model’s general language capabilities.\\
In addition to decoder-based models, we evaluate the effectiveness of encoder-only architectures. We use \texttt{RoBERTa-base} \cite{liu2019roberta} as a representative encoder model. Due to its limited input length, we adopt a two-step chunking and aggregation pipeline: input transcripts are first segmented into overlapping chunks, encoded individually, and then aggregated via mean pooling to generate a global representation. A lightweight nonlinear regression head is trained on these pooled embeddings to predict trait scores.
\paragraph{Embedding-based Regression}
To complement prompt- and fine-tuning-based approaches, we evaluate static embedding-based personality inference using two pretrained sentence encoders. First, we adopt \texttt{all-MiniLM-L6-v2}, a compact and efficient transformer model from the Sentence-Transformers library~\cite{allMiniLM2021}, built upon the Sentence-BERT framework~\cite{reimers2019sentencebert}. This model maps variable-length interview responses into fixed-size dense representations optimized for semantic similarity tasks.
Second, we utilize OpenAI’s \texttt{text-embedding-3-small}~\cite{openai2024embedding}, accessed via the OpenAI API (model ID: \texttt{text-embedding-3-small}, version: 2024-04-15), accessed via the OpenAI API, which generates 1,536-dimensional embeddings designed for general-purpose applications such as retrieval, clustering, and classification. For both models, we compute embeddings at the full-transcript level and use them as input to a multi-output Ridge regression model to predict the five Big Five trait scores. All input embeddings are standardized, and the dataset is split into 80\% training and 20\% testing partitions..
\paragraph{Evaluation Metrics}
We assess model performance using two complementary metrics: Pearson correlation ($r$) and Mean Absolute Error (MAE). Pearson correlation quantifies the consistency between model-predicted and ground-truth trait rankings, serving as a proxy for criterion validity. Correlations above $r = 0.5$ are typically interpreted as strong evidence of convergent validity, indicating that the model captures meaningful individual differences in personality. Values between $0.3$ and $0.5$ suggest moderate alignment, while correlations below $0.3$ imply weak predictive consistency.\\
MAE provides a direct measure of prediction error on the original trait scale (1.0–5.0). Lower MAE values (e.g., $\leq 0.5$) reflect high precision, while values between $0.5$ and $0.8$ are common and may be acceptable depending on context. MAE values exceeding $1.0$ typically signal poor calibration or unreliable estimates. Unlike correlation, MAE is sensitive to systematic bias and offers a more interpretable estimate of absolute prediction fidelity.

\section{Results}
\begin{figure}[ht]
    \centering
    \includegraphics[width=0.9\linewidth]{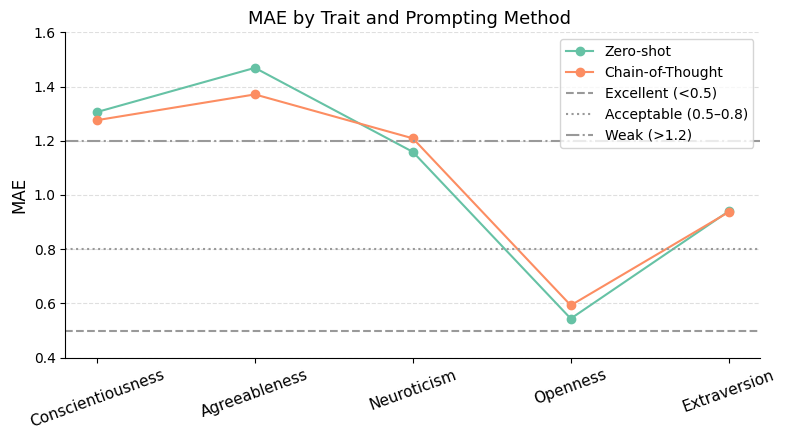}
    \caption{ MAE of GPT-4.1 Mini Across Big Five Traits by Prompting Method.}
    \label{fig:mae_plot}
    \vspace{-1em}
\end{figure}

\paragraph{Prompt-based Inference Results}
Table~\ref{tab:gpt_r_only} summarizes the of performance of GPT-4.1 Mini using zero-shot and chain-of-thought prompting. The strongest alignment was observed for Conscientiousness in the zero-shot setting ($r = 0.250$), followed by modest correlation on Agreeableness ($r = 0.132$). For the remaining traits, Neuroticism, Openness, and Extraversion showed near-zero or negative correlations, indicating limited predictive validity.
\begin{table}[t]
\centering
\caption{Pearson correlation ($r$) of GPT-4.1 Mini on Big Five trait prediction}
\begin{tabular}{lcc}
\toprule
\textbf{Trait} & \textbf{Zero-shot} & \textbf{Chain-of-Thought} \\
\midrule
Conscientiousness & \textbf{0.250} & 0.236 \\
Agreeableness     & 0.132 & \textbf{0.133} \\
Neuroticism       & \textbf{0.065} & $-$0.009 \\
Openness          & 0.020 & \textbf{0.051} \\
Extraversion      & \textbf{0.041} & 0.005 \\
\bottomrule
\end{tabular}
\label{tab:gpt_r_only}
\end{table}
\\
Unexpectedly, chain-of-thought prompting did not improve performance and in several cases led to slight degradation. Across all five traits, prediction errors remained high. Only Openness achieved a mean absolute error (MAE) below 0.6, while Conscientiousness and Agreeableness exceeded 1.2, indicating poor calibration. As visualized in Figure~\ref{fig:mae_plot}, gains from CoT prompting were inconsistent and marginal.
\paragraph{LoRA-based Adaptation Results}
Table~\ref{tab:lora_r} presents the Pearson correlation coefficients between predicted and ground-truth Big Five trait scores across two LoRA-based modeling strategies. Across all traits, the embedding-based models consistently outperformed their Llama-based counterparts. In particular, Neuroticism and Openness achieved the most robust alignment within the embedding+LoRA setup, with $r > 0.20$ across all ranks. The highest overall correlation was observed for Openness ($r = 0.255$) at Rank 32. In contrast, the Llama-LoRA models showed modest performance, with Conscientiousness achieving the best result ($r = 0.164$ at Rank 8), while other traits, especially Extraversion and Openness exhibited weak or negative correlations.
\begin{table}[H]
\centering
\small
\renewcommand{\arraystretch}{1.5}
\setlength{\tabcolsep}{4pt}
\begin{threeparttable}
\caption{Pearson $r$ correlations across Big Five traits using LoRA fine-tuning.}
\label{tab:lora_r}
\begin{tabular}{lccccc}
\toprule
\textbf{Model (LoRA)} & \textbf{E} & \textbf{A} & \textbf{C} & \textbf{N} & \textbf{O} \\
\midrule
Rank 8 (BERT)  & 0.135 & 0.045 & \textbf{0.219} & 0.202 & 0.198 \\
Rank 16 (BERT) & 0.136 & 0.172 & 0.143 & 0.232 & 0.251 \\
\rowcolor{pink!20} Rank 32 (BERT) & \textbf{0.197} & \textbf{0.173} & -0.098 & \textbf{0.236} & \textbf{0.255} \\
Rank 8 (Llama )  & -0.128 & 0.096 & 0.164 & 0.088 & -0.064 \\
Rank 16 (Llama ) & -0.027 & -0.065 & -0.025 & 0.027 & 0.051 \\
Rank 32 (Llama ) & -0.060 & 0.078 & -0.050 & 0.108 & -0.095 \\
\bottomrule
\end{tabular}
\begin{tablenotes}
\footnotesize
\item \textbf{Note.} Big Five traits: E = Extraversion, A = Agreeableness, C = Conscientiousness, N = Neuroticism, O = Openness. 
\end{tablenotes}
\end{threeparttable}
\end{table}
 
\begin{figure}[ht]
    \centering
    \includegraphics[width=\linewidth]{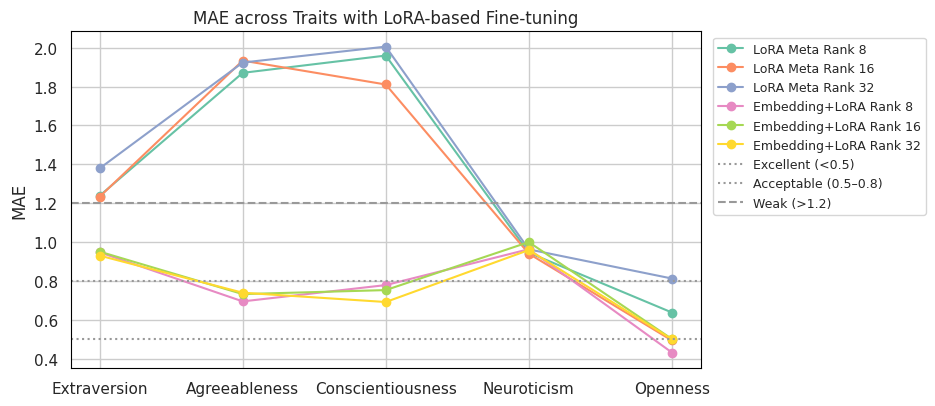}
    \caption{MAE across Big Five traits for LoRA fine-tuning methods.}
    \label{fig:lora_mae}
    \end{figure}

Figure~\ref{fig:lora_mae} presents the Mean Absolute Error (MAE) across all five Big Five personality traits, providing a direct quantification of prediction error on the original rating scale. 
The embedding-based models (BERT + LoRA) consistently achieved lower MAE scores compared to the Meta-Llama LoRA variants. Across all traits, none of the embedding-based models exceeded MAE = 1.0, indicating generally reliable alignment. In contrast, all Meta-Llama variants produced MAE greater than 1.2 for Extraversion, Agreeableness, and Conscientiousness, with the only exception being Openness, which achieved its lowest MAE of 0.431 at Rank 8. Openness was also the most precisely predicted trait in the Meta-Llama group, with its best result occurring at Rank 16 (MAE = 0.493).
Within the BERT embedding-based group, Extraversion and Neuroticism remained the most challenging traits, yet the best results—MAE = 0.930 and 0.691 respectively—were observed at Rank 32. For Agreeableness, the best MAE (0.694) was obtained at Rank 8, while Conscientiousness reached its lowest MAE (0.691) at Rank 32, both falling within the acceptable range (0.5–0.8). 

\begin{figure}[ht]
    \centering
    \includegraphics[width=0.9\linewidth]{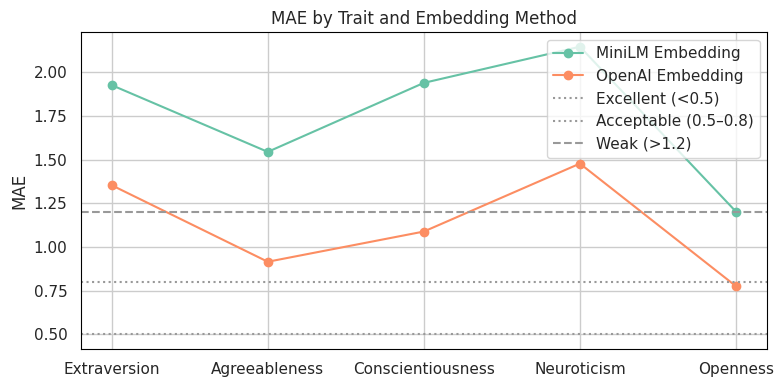}
    \caption{MAE across personality traits using MiniLM and OpenAI embeddings. Colored reference lines denote qualitative thresholds.}
    \label{fig:embedding_mae_plot}
\end{figure}

\paragraph{Embedding-based Regression Results}
To complement instruction-tuned LLMs, we also evaluated static sentence embedding models as trait predictors. Specifically, we used two pre-trained encoders: all-MiniLM-L6-v2 from SentenceTransformers and OpenAI’s text-embedding-3-small. Table~\ref{tab:embedding_r} displays the Pearson correlation ($r$) and Figure~\ref{fig:embedding_mae_plot} represent the mean absolute error (MAE) for each trait. \\

\begin{table}[ht]
\centering
\caption{Pearson correlation ($r$) for each trait using MiniLM and OpenAI embeddings.}
\begin{tabular}{lcc}
\toprule
\textbf{Trait} & \textbf{MiniLM} & \textbf{OpenAI} \\
\midrule
Extraversion      & 0.048 & \textbf{0.086} \\
Agreeableness     & -0.069 & \textbf{0.119} \\
Conscientiousness & \textbf{0.106} & 0.070 \\
Neuroticism       & \textbf{0.112} & -0.050 \\
Openness          & 0.025 & \textbf{0.050} \\
\bottomrule
\end{tabular}
\label{tab:embedding_r}
\end{table}
Overall, both embedding-based approaches showed limited alignment with ground-truth personality traits. Pearson $r$ values remained low across all five traits, with all scores below 0.12. The best result for MiniLM was on Neuroticism ($r$ = 0.112), while OpenAI embeddings achieved the highest correlation on Agreeableness ($r$ = 0.119). Notably, Agreeableness was also the worst-performing trait for MiniLM ($r$ = –0.069), and Neuroticism the worst for OpenAI embeddings ($r$ = –0.050), indicating divergent alignment patterns between the two encoders. 

In terms of prediction error (MAE), both models exhibited relatively poor calibration. Most MAE values exceeded 1.0, with OpenAI embeddings generally outperforming MiniLM. The best result was achieved on Openness (MAE = 0.776), followed by Agreeableness (MAE = 0.916). Conscientiousness reached MAE = 1.080, while both Extraversion and Neuroticism exceeded 1.2, suggesting coarse trait estimation.

\section{Discussion and Conclusion}
This study set out to evaluate the alignment capacity of large language models (LLMs) and related methods in predicting continuous Big Five personality trait scores from real-world conversation data. Personality prediction from natural language poses a unique alignment challenge: it requires not only semantic understanding, but also subtle inference over emotion, intention, and long-term behavioral patterns—all projected onto a continuous 1–5 Likert scale. We benchmarked four paradigms: zero-shot and chain-of-thought prompting with GPT-4, LoRA-based fine-tuning on Meta-LLaMA and RoBERTa, static sentence embedding models, and hybrid embedding+LoRA regressors. Across these methods, we observed limited alignment with ground-truth trait scores: most Pearson correlations remained below 0.3, with several falling near zero or negative. Complementarily, MAE scores commonly exceeded 1.0, indicating that models often deviated by more than one full Likert scale point from true values—far from acceptable precision for personality-related applications. While a few models achieved MAE values near 0.5 on individual traits (e.g., Openness or Conscientiousness), such cases were rare and inconsistent. These results highlight the overall difficulty of aligning LLM behavior with high-resolution, psychologically grounded constructs such as personality.\\
The relative trait-level performance trends further illuminate how different modeling paradigms struggle with different facets of personality. GPT-4.1 performed comparatively better on Conscientiousness ($r \approx 0.25$) and Agreeableness ($r \approx 0.13$), two traits associated with rule-following, goal-directed behaviors, and social harmony. This may reflect GPT's alignment with normative patterns and instruction-following behavior learned during RLHF training. In contrast, LoRA fine-tuned encoder-based models (e.g., RoBERTa) exhibited more consistent improvements in Openness and Neuroticism as LoRA rank increased. These traits, linked to imagination and emotional reactivity respectively, may be more dependent on sentiment comprehension and internal affect representation—domains where encoder-based embeddings, combined with task-specific adaptation, excel.\\
Interestingly, Openness, which emphasizes creativity, fantasy, and abstract thinking, was better captured by auto-regressive models (e.g., LLaMA) at lower MAE, even when their Pearson r was lower. This divergence between correlation and absolute error highlights a core challenge in alignment: achieving both calibration and consistency in real-valued, psychologically grounded outputs. Conversely, Extraversion, inherently linked to social expressiveness and interpersonal energy was poorly predicted across all methods. This may be due to the single-speaker format of the input interviews, which lack the interactive, behavioral cues necessary for reliable modeling of sociability or assertiveness.\\
These findings underscore key implications for the development and deployment of aligned AI systems in human-centered domains. First, the generally low trait alignment suggests that current LLMs, even when fine-tuned or augmented, lack the precision and interpretability necessary for reliable personality inference. This limitation is particularly concerning for traits such as Neuroticism, which correlate with mental health risks and emotional vulnerability, where prediction errors could lead to harmful downstream outcomes. Second, our results suggest that trait-specific modeling challenges reflect fundamental differences in how various traits are expressed in language: Openness may be tied to imaginative expression that decoders like LLaMA struggle to internalize without explicit guidance, whereas Conscientiousness may be more observable through structured linguistic behaviors captured better by prompt-tuned or encoder-based models. Third, these challenges reveal that alignment in the personality domain requires not only architectural adjustments but also deeper modeling of context, self-reflection, and longitudinal cues, elements that go beyond single-pass text prediction.\\

\paragraph{Limitation and Future Work}
While this study provides a comprehensive comparison of alignment strategies for personality prediction, several limitations constrain the generalizability and scope of our findings. First, the data set used, although derived from real-world conversation, is modest in size and may not fully support the parameter needs of large-scale fine tuning, particularly for decoder-based LLMs such as Meta-LLaMA. Additionally, the length of conversational input may affect model outputs, especially in embedding-chunk-based architectures. Future work should explore more sophisticated chunking algorithms and hierarchical representations that allow smaller models to better encode long-range context.\\
Second, our use of GPT-based prompting was limited to zero-shot and chain-of-thought (CoT) strategies. While these are meaningful baselines, they do not fully reflect the prompt engineering space. Exploring few-shot prompts and formats that explicitly integrate emotional or contextual cues may enhance model performance, particularly on traits like Consciousness, where surface linguistic features alone may be insufficient for inference.
Third, we varied LoRA rank during fine-tuning, but did not explore other hyperparameters such as temperature scaling, dropout strategies, or layer-wise adaptation. Traits such as \textit{Neuroticism}, which involve emotional instability and are linked to mental health vulnerability, may require more tailored fine-tuning regimes for reliable alignment. Improved modeling of these traits could have significant implications for early-stage psychological screening and support.\\
Fourth, the current evaluation was limited to a core set of widely used general-purpose models. Although this provides a solid foundation for comparative analysis, future work should extend to domain-adapted models (e.g., mental health–oriented LLMs) and investigate how extract specific features affect alignment quality.
Fifth, our current setup predicts all five personality dimensions simultaneously using a single model prompt. This multi-target formulation may dilute the model’s attention and lead to less focused predictions. A trait-specific prompting strategy with tailored input patterns and conditioning for each dimension may improve the alignment, particularly for traits requiring distinct cognitive or emotional inference pathways.
Finally, personality traits such as \textit{Agreeableness}, which involve interpersonal behaviors like empathy and cooperativeness, may not be fully inferable from textual content alone. These constructs often manifest through observable social interactions and event-level context. Future work could integrate behavioral event extraction, structured memory, or multimodal input (e.g., facial expression, interaction patterns) to improve trait-specific grounding.
\paragraph{Conclusion}
We evaluated a range of large language model approaches for predicting Big Five personality traits from conversational data, highlighting challenges in alignment with psychologically grounded targets. Across prompting, fine-tuning, and embedding-based methods, performance remained limited, with low trait-level correlations and high calibration error. Our findings reveal that personality assessment poses a uniquely demanding alignment task, particularly for traits tied to emotion, cognition, and social behavior. \\
These results underscore the inherent difficulty of inferring latent personality constructs from unstructured text. To address this, we advocate for future work along three directions: (1) developing more targeted prompt engineering strategies, including trait-specific instructions, self-consistency, and retrieval-augmented prompting; (2) incorporating long-range autobiographical memory modeling to capture identity-related temporal context; and (3) extending beyond text to multimodal signals such as speech prosody, facial expressions, and interaction dynamics. Bridging these advances may enable more robust, ethically sound personality inference for real-world, human-centered AI systems.

\section{Ethical Considerations}
While LLMs offer scalable potential for personality assessment, their use raises concerns around privacy, consent, and interpretability. We caution against deploying these models in high-stakes settings without human oversight and emphasize the need for theory-grounded, transparent evaluation protocols.


\bibliography{aaai2026}

\begin{thebibliography}{49}
\providecommand{\natexlab}[1]{#1}

\bibitem[{AI(2025)}]{llama32025}
AI, M. 2025.
\newblock Meta Llama 3.1 Collection.
\newblock \url{https://huggingface.co/meta-llama/Llama-3.1-8B-Instruct}.
\newblock Accessed: 2025-08-01.

\bibitem[{Allport and Odbert(1936)}]{allport1936trait}
Allport, G.; and Odbert, H. 1936.
\newblock \emph{Trait-Names: A Psycho-lexical Study. No. 211.}
\newblock Princeton.

\bibitem[{Allport(1937)}]{allport1937personality}
Allport, G.~W. 1937.
\newblock \emph{Personality: A Psychological Interpretation}.
\newblock Holt.

\bibitem[{Bleidorn et~al.(2019)Bleidorn, Hill, Back, Denissen, Hennecke,
  Hopwood, Jokela, Kandler, Lucas, Luhmann et~al.}]{bleidorn2019policy}
Bleidorn, W.; Hill, P.~L.; Back, M.~D.; Denissen, J.~J.; Hennecke, M.; Hopwood,
  C.~J.; Jokela, M.; Kandler, C.; Lucas, R.~E.; Luhmann, M.; et~al. 2019.
\newblock The policy relevance of personality traits.
\newblock \emph{American psychologist}, 74(9): 1056.

\bibitem[{Caldwell and Burger(1998)}]{caldwell1998personality}
Caldwell, D.~F.; and Burger, J.~M. 1998.
\newblock Personality characteristics of job applicants and success in
  screening interviews.
\newblock \emph{Personnel Psychology}, 51(1): 119--136.

\bibitem[{Christian et~al.(2021)Christian, Suhartono, Chowanda, and
  Zamli}]{christian2021text}
Christian, H.; Suhartono, D.; Chowanda, A.; and Zamli, K.~Z. 2021.
\newblock Text based personality prediction from multiple social media data
  sources using pre-trained language model and model averaging.
\newblock \emph{Journal of Big Data}, 8(1): 68.

\bibitem[{Coifman and Bonanno(2010)}]{coifman2010distress}
Coifman, K.; and Bonanno, G. 2010.
\newblock When distress does not become depression: Emotion context sensitivity
  and adjustment to bereavement.
\newblock \emph{Journal of Abnormal Psychology}, 119(3): 479--490.

\bibitem[{Coifman et~al.(2007)Coifman, Bonanno, Ray, and
  Gross}]{coifman2007repressive}
Coifman, K.; Bonanno, G.; Ray, R.; and Gross, J. 2007.
\newblock Does repressive coping promote resilience? Affective-autonomic
  response discrepancy during bereavement.
\newblock \emph{Journal of Personality and Social Psychology}, 92(4): 745--758.

\bibitem[{Coifman, Flynn, and Pinto(2016)}]{coifman2016context}
Coifman, K.; Flynn, J.; and Pinto, L. 2016.
\newblock When context matters: Negative emotions predict psychological health
  and adjustment.
\newblock \emph{Motivation \& Emotion}, 40(4): 602--624.

\bibitem[{de~Sousa and Sequeira(2025)}]{de2025personality}
de~Sousa, L.~G.; and Sequeira, J.~S. 2025.
\newblock Personality Trait Prediction Using Text.
\newblock In \emph{Robot Intelligence Technology and Applications 9: Results
  from the 12th International Conference on Robot Intelligence Technology and
  Applications}, volume 1419, 332. Springer Nature.

\bibitem[{Devlin et~al.(2019)Devlin, Chang, Lee, and
  Toutanova}]{devlin2019bert}
Devlin, J.; Chang, M.-W.; Lee, K.; and Toutanova, K. 2019.
\newblock Bert: Pre-training of deep bidirectional transformers for language
  understanding.
\newblock In \emph{Proceedings of the 2019 conference of the North American
  chapter of the association for computational linguistics: human language
  technologies, volume 1 (long and short papers)}, 4171--4186.

\bibitem[{Gjurkovi{\'c} et~al.(2020)Gjurkovi{\'c}, Karan, Vukojevi{\'c},
  Bo{\v{s}}njak, and {\v{S}}najder}]{gjurkovic2020pandora}
Gjurkovi{\'c}, M.; Karan, M.; Vukojevi{\'c}, I.; Bo{\v{s}}njak, M.; and
  {\v{S}}najder, J. 2020.
\newblock PANDORA talks: Personality and demographics on Reddit.
\newblock \emph{arXiv preprint arXiv:2004.04460}.

\bibitem[{Harvey et~al.(2014)Harvey, Coifman, Ross, Kleinert, and
  Giardina}]{harvey2014emotion}
Harvey, M.; Coifman, K.; Ross, G.; Kleinert, D.; and Giardina, P. 2014.
\newblock Contextually appropriate emotion-word use predicts adaptive health
  behavior: Emotion context sensitivity and treatment adherence.
\newblock \emph{Journal of Health Psychology}.
\newblock Advance online publication.

\bibitem[{Hu et~al.(2022)Hu, Shen, Wallis, Allen-Zhu, Li, Wang, Wang, Chen
  et~al.}]{hu2022lora}
Hu, E.~J.; Shen, Y.; Wallis, P.; Allen-Zhu, Z.; Li, Y.; Wang, S.; Wang, L.;
  Chen, W.; et~al. 2022.
\newblock Lora: Low-rank adaptation of large language models.
\newblock \emph{ICLR}, 1(2): 3.

\bibitem[{Hu et~al.(2024)Hu, He, Wang, Zhao, Shao, and Nie}]{hu2024llm}
Hu, L.; He, H.; Wang, D.; Zhao, Z.; Shao, Y.; and Nie, L. 2024.
\newblock Llm vs small model? large language model based text augmentation
  enhanced personality detection model.
\newblock In \emph{Proceedings of the AAAI Conference on Artificial
  Intelligence}, volume~38, 18234--18242.

\bibitem[{Huang et~al.(2024)Huang, Zhang, Soto, and Evans}]{huang2024designing}
Huang, M.; Zhang, X.; Soto, C.; and Evans, J. 2024.
\newblock Designing llm-agents with personalities: A psychometric approach.
\newblock \emph{arXiv preprint arXiv:2410.19238}.

\bibitem[{Ji et~al.(2023)Ji, Wu, Zheng, Hu, Chen, and He}]{ji2023chatgpt}
Ji, Y.; Wu, W.; Zheng, H.; Hu, Y.; Chen, X.; and He, L. 2023.
\newblock Is chatgpt a good personality recognizer? a preliminary study.
\newblock \emph{arXiv preprint arXiv:2307.03952}.

\bibitem[{Jiang et~al.(2023{\natexlab{a}})Jiang, Xu, Zhu, Han, Zhang, and
  Zhu}]{jiang2023evaluating}
Jiang, G.; Xu, M.; Zhu, S.-C.; Han, W.; Zhang, C.; and Zhu, Y.
  2023{\natexlab{a}}.
\newblock Evaluating and inducing personality in pre-trained language models.
\newblock \emph{Advances in Neural Information Processing Systems}, 36:
  10622--10643.

\bibitem[{Jiang et~al.(2023{\natexlab{b}})Jiang, Zhang, Cao, Breazeal, Roy, and
  Kabbara}]{hang_jiang2023personallm}
Jiang, H.; Zhang, X.; Cao, X.; Breazeal, C.; Roy, D.; and Kabbara, J.
  2023{\natexlab{b}}.
\newblock PersonaLLM: Investigating the ability of large language models to
  express personality traits.
\newblock \emph{arXiv preprint arXiv:2305.02547}.

\bibitem[{John, Srivastava et~al.(1999)}]{john1999big}
John, O.~P.; Srivastava, S.; et~al. 1999.
\newblock \emph{The Big-Five trait taxonomy: History, measurement, and
  theoretical perspectives}.
\newblock University of California Berkeley.

\bibitem[{Kamalesh and Bharathi(2022)}]{kamalesh2022personality}
Kamalesh, M.~D.; and Bharathi, B. 2022.
\newblock Personality prediction model for social media using machine learning
  Technique.
\newblock \emph{Computers and Electrical Engineering}, 100: 107852.

\bibitem[{Kojima et~al.(2023)Kojima, Gu, Reid, Matsuo, and
  Iwasawa}]{kojima2023largelanguagemodelszeroshot}
Kojima, T.; Gu, S.~S.; Reid, M.; Matsuo, Y.; and Iwasawa, Y. 2023.
\newblock Large Language Models are Zero-Shot Reasoners.
\newblock arXiv:2205.11916.

\bibitem[{Lewis et~al.(2020)Lewis, Perez, Piktus, Petroni, Karpukhin, Goyal,
  K{\"u}ttler, Lewis, Yih, Rockt{\"a}schel et~al.}]{lewis2020retrieval}
Lewis, P.; Perez, E.; Piktus, A.; Petroni, F.; Karpukhin, V.; Goyal, N.;
  K{\"u}ttler, H.; Lewis, M.; Yih, W.-t.; Rockt{\"a}schel, T.; et~al. 2020.
\newblock Retrieval-augmented generation for knowledge-intensive nlp tasks.
\newblock \emph{Advances in neural information processing systems}, 33:
  9459--9474.

\bibitem[{Liu et~al.(2019)Liu, Ott, Goyal, Du, Joshi, Chen, Levy, Lewis,
  Zettlemoyer, and Stoyanov}]{liu2019roberta}
Liu, Y.; Ott, M.; Goyal, N.; Du, J.; Joshi, M.; Chen, D.; Levy, O.; Lewis, M.;
  Zettlemoyer, L.; and Stoyanov, V. 2019.
\newblock Roberta: A robustly optimized bert pretraining approach.
\newblock \emph{arXiv preprint arXiv:1907.11692}.

\bibitem[{Maharjan et~al.(2025{\natexlab{a}})Maharjan, Jin, Zhu, and
  Kenne}]{julina_personality_su}
Maharjan, J.; Jin, R.; Zhu, J.; and Kenne, D. 2025{\natexlab{a}}.
\newblock Intersection of Big Five Personality Traits and Substance Use on X:
  Insight from the COVID-19 Pandemic.
\newblock \emph{Journal of Medical Internet Research}.

\bibitem[{Maharjan et~al.(2025{\natexlab{b}})Maharjan, Jin, Zhu, and
  Kenne}]{maharjan2025psychometric}
Maharjan, J.; Jin, R.; Zhu, J.; and Kenne, D. 2025{\natexlab{b}}.
\newblock Psychometric Evaluation of Large Language Model Embeddings for
  Personality Trait Prediction.
\newblock \emph{Journal of Medical Internet Research}, 27: e75347.

\bibitem[{Maharjan et~al.(2025{\natexlab{c}})Maharjan, Zhu, King, Phan, Kenne,
  and Jin}]{maharjan2025large}
Maharjan, J.; Zhu, J.; King, J.; Phan, N.; Kenne, D.; and Jin, R.
  2025{\natexlab{c}}.
\newblock Large-Scale Deep Learning--Enabled Infodemiological Analysis of
  Substance Use Patterns on Social Media: Insights From the COVID-19 Pandemic.
\newblock \emph{JMIR Infodemiology}, 5: e59076.

\bibitem[{{Mistral AI}(2024)}]{mistral2024}
{Mistral AI}. 2024.
\newblock \emph{Mistral 7B v0.2 Instruct}.
\newblock Model release. Version 0.2 of Mistral-7B with instruction tuning.

\bibitem[{Mohammad and Turney(2013)}]{mohammad2013nrc}
Mohammad, S.~M.; and Turney, P.~D. 2013.
\newblock Nrc emotion lexicon.
\newblock \emph{National Research Council, Canada}, 2: 234.

\bibitem[{Murray(1936)}]{murray1936basic}
Murray, H.~A. 1936.
\newblock Basic concepts for a psychology of personality.
\newblock \emph{The Journal of general psychology}, 15(2): 241--268.

\bibitem[{OpenAI(2024)}]{openai2024embedding}
OpenAI. 2024.
\newblock OpenAI Text Embeddings.
\newblock \url{https://platform.openai.com/docs/guides/embeddings}.
\newblock Accessed: 2025-08-01.

\bibitem[{OpenAI(2025)}]{openai2025gpt41mini}
OpenAI. 2025.
\newblock GPT-4.1 Mini.
\newblock \url{https://platform.openai.com/docs/models/gpt-4.1-mini}.
\newblock Accessed: 2025-08-01.

\bibitem[{Park et~al.(2023)Park, O'Brien, Cai, Morris, Liang, and
  Bernstein}]{park2023generative}
Park, J.~S.; O'Brien, J.; Cai, C.~J.; Morris, M.~R.; Liang, P.; and Bernstein,
  M.~S. 2023.
\newblock Generative agents: Interactive simulacra of human behavior.
\newblock In \emph{Proceedings of the 36th annual acm symposium on user
  interface software and technology}, 1--22.

\bibitem[{Pennebaker et~al.(2001)Pennebaker, Francis, Booth
  et~al.}]{pennebaker2001linguistic}
Pennebaker, J.~W.; Francis, M.~E.; Booth, R.~J.; et~al. 2001.
\newblock Linguistic inquiry and word count: LIWC 2001.
\newblock \emph{Mahway: Lawrence Erlbaum Associates}, 71(2001): 2001.

\bibitem[{Peters and Matz(2024)}]{peters2024large}
Peters, H.; and Matz, S.~C. 2024.
\newblock Large language models can infer psychological dispositions of social
  media users.
\newblock \emph{PNAS nexus}, 3(6): pgae231.

\bibitem[{Reimers and Gurevych(2019)}]{reimers2019sentencebert}
Reimers, N.; and Gurevych, I. 2019.
\newblock Sentence-BERT: Sentence Embeddings using Siamese BERT-Networks.
\newblock In \emph{Proceedings of the 2019 Conference on Empirical Methods in
  Natural Language Processing}.

\bibitem[{Reimers and Gurevych(2021)}]{allMiniLM2021}
Reimers, N.; and Gurevych, I. 2021.
\newblock all-MiniLM-L6-v2: A Sentence-Transformer Model.
\newblock \url{https://huggingface.co/sentence-transformers/all-MiniLM-L6-v2}.
\newblock Accessed: 2025-08-01.

\bibitem[{Renze(2024)}]{Renze_2024}
Renze, M. 2024.
\newblock The Effect of Sampling Temperature on Problem Solving in Large
  Language Models.
\newblock In \emph{Findings of the Association for Computational Linguistics:
  EMNLP 2024}, 7346–7356. Association for Computational Linguistics.

\bibitem[{Schwartz et~al.(2013)Schwartz, Eichstaedt, Kern, Dziurzynski,
  Ramones, Agrawal, Shah, Kosinski, Stillwell, Seligman
  et~al.}]{schwartz2013personality}
Schwartz, H.~A.; Eichstaedt, J.~C.; Kern, M.~L.; Dziurzynski, L.; Ramones,
  S.~M.; Agrawal, M.; Shah, A.; Kosinski, M.; Stillwell, D.; Seligman, M.~E.;
  et~al. 2013.
\newblock Personality, gender, and age in the language of social media: The
  open-vocabulary approach.
\newblock \emph{PloS one}, 8(9): e73791.

\bibitem[{Serapio-Garc{\'\i}a et~al.(2023)Serapio-Garc{\'\i}a, Safdari, Crepy,
  Sun, Fitz, Abdulhai, Faust, and Matari{\'c}}]{serapio2023personality}
Serapio-Garc{\'\i}a, G.; Safdari, M.; Crepy, C.; Sun, L.; Fitz, S.; Abdulhai,
  M.; Faust, A.; and Matari{\'c}, M. 2023.
\newblock Personality traits in large language models.

\bibitem[{Sirasapalli and Malla(2023)}]{sirasapalli2023deep}
Sirasapalli, J.~J.; and Malla, R.~M. 2023.
\newblock A deep learning approach to text-based personality prediction using
  multiple data sources mapping.
\newblock \emph{Neural Computing and Applications}, 35(28): 20619--20630.

\bibitem[{Soto and John(2017)}]{bfi2}
Soto, C.~J.; and John, O.~P. 2017.
\newblock The next Big Five Inventory (BFI-2): Developing and assessing a
  hierarchical model with 15 facets to enhance bandwidth, fidelity, and
  predictive power.
\newblock \emph{Journal of personality and social psychology}, 113(1): 117.

\bibitem[{Touvron et~al.(2023)Touvron, Lavril, Izacard, Martinet, Lachaux,
  Lacroix, Rozi{\`e}re, Goyal, Hambro, Azhar, Rodriguez, Joulin, Grave, and
  Lample}]{touvron2023llama}
Touvron, H.; Lavril, T.; Izacard, G.; Martinet, X.; Lachaux, M.-A.; Lacroix,
  T.; Rozi{\`e}re, B.; Goyal, N.; Hambro, E.; Azhar, F.; Rodriguez, A.; Joulin,
  A.; Grave, E.; and Lample, G. 2023.
\newblock LLaMA: Open and Efficient Foundation Language Models.
\newblock \emph{arXiv preprint}.
\newblock ArXiv:2302.13971.

\bibitem[{Trull and Durrett(2005)}]{trull2005categorical}
Trull, T.~J.; and Durrett, C.~A. 2005.
\newblock Categorical and dimensional models of personality disorder.
\newblock \emph{Annu. Rev. Clin. Psychol.}, 1(1): 355--380.

\bibitem[{Wang et~al.(2025)Wang, Zhao, Ones, He, and Xu}]{wang2025evaluating}
Wang, Y.; Zhao, J.; Ones, D.~S.; He, L.; and Xu, X. 2025.
\newblock Evaluating the ability of large language models to emulate
  personality.
\newblock \emph{Scientific reports}, 15(1): 519.

\bibitem[{Wang et~al.(2023)Wang, Peng, Que, Liu, Zhou, Wu, Guo, Gan, Ni, Yang
  et~al.}]{wang2023rolellm}
Wang, Z.~M.; Peng, Z.; Que, H.; Liu, J.; Zhou, W.; Wu, Y.; Guo, H.; Gan, R.;
  Ni, Z.; Yang, J.; et~al. 2023.
\newblock Rolellm: Benchmarking, eliciting, and enhancing role-playing
  abilities of large language models.
\newblock \emph{arXiv preprint arXiv:2310.00746}.

\bibitem[{Wei et~al.(2023)Wei, Wang, Schuurmans, Bosma, Ichter, Xia, Chi, Le,
  and Zhou}]{wei2023chainofthoughtpromptingelicitsreasoning}
Wei, J.; Wang, X.; Schuurmans, D.; Bosma, M.; Ichter, B.; Xia, F.; Chi, E.; Le,
  Q.; and Zhou, D. 2023.
\newblock Chain-of-Thought Prompting Elicits Reasoning in Large Language
  Models.
\newblock arXiv:2201.11903.

\bibitem[{Zhu, Jin, and Coifman(2025)}]{zhu2025llmsinferpersonalityreal}
Zhu, J.; Jin, R.; and Coifman, K.~G. 2025.
\newblock Can LLMs Infer Personality from Real World Conversations?
\newblock arXiv:2507.14355.

\bibitem[{Zhu et~al.(2025)Zhu, Zhang, Jin, Jiang, and Kenne}]{zhu2025probing}
Zhu, J.; Zhang, X.; Jin, R.; Jiang, H.; and Kenne, D.~R. 2025.
\newblock Probing Public Perceptions of Antidepressants on Social Media: Mixed
  Methods Study.
\newblock \emph{JMIR Formative Research}, 9(1): e62680.

\end{thebibliography}

\end{document}